\documentclass[10pt,twocolumn,letterpaper]{article}

\usepackage[pagenumbers]{wacv} 
\usepackage{algorithm}
\usepackage{algpseudocode}
\usepackage{amsmath, amssymb}
\usepackage{array}
\usepackage{multirow}
\usepackage{float}
\usepackage{subcaption}

\definecolor{wacvblue}{rgb}{0.21,0.49,0.74}
\usepackage[pagebackref,breaklinks,colorlinks,allcolors=wacvblue]{hyperref}

\def\wacvPaperID{967} 
\def\confName{WACV}
\def\confYear{2027}

\title{MoCKEFA: Model-Conditioned Keyframe Extraction and Fair Attribution for Video Model Explanation}

\author{Lahiru Wijayasingha\\
Independent Researcher\\
{\tt\small xlahirunuwan@gmail.com}\\
{\tt\small \url{https://cyclopesvision.com}}
}

\begin{document}
\maketitle
\begin{abstract}
We show that a video can be represented with just a few keyframes without sacrificing a given model's belief about the video. Our method is the first in it's kind that extracts these keyframes conditioned on the model's prediction which is massively helpful for building explainable video models. We then fairly distribute the importance of the video to the model among the keyframes with game theoretically fair Shapley values. Through qualitative and quantitative experiments, we show that keyframe attribution is superior to the traditional frame attribution. Further we conduct preliminary exploration on highlighting the important changes among the frames by looking at the differences between adjacent keyframes.    
\end{abstract}
    
\section{Introduction}
\label{sec:intro}

Temporal attribution quantifies each frame's influence on a video model's decision. However, some adjacent frames are highly correlated \cite{goran2020adjacent}\cite{pan2016learning} and some are noisy making per-frame analysis redundant and difficult to interpret. Our algorithm asks the model itself; not any external entity "which frames are actually important for your decision ?". The answer to this lets us represent the whole video with a few keyframes that the model values. Then we judge the relative importance of these keyframes by "fairly" distributing the total importance of the video among the keyframes. This shift not only enhances interpretability but also significantly reduces computational overhead.

\begin{figure}[t]
	\centering
	\includegraphics[width=0.9\linewidth]{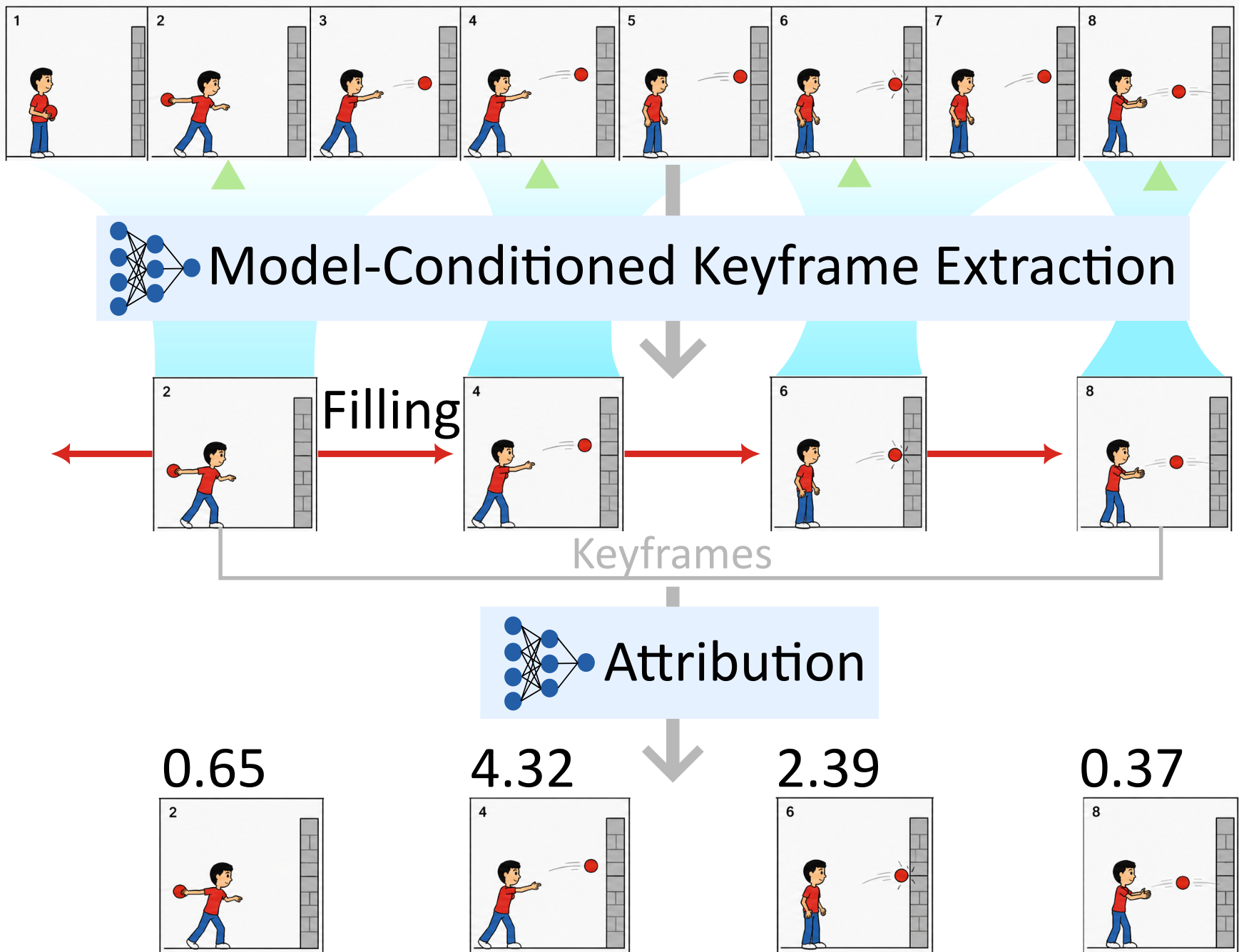}
	\caption{MoCKEFA framework. Keyframe extraction: model selects keyframes that are important for model's decision. Continuous temporal regions can be represented with just a selected keyframe with minimal change to the model belief. Fair attribution: Shapley values are calculated for each keyframe that gives game-theoretically "fair" attributions}
	\label{fig:intro}
\end{figure}

Any method that interprets a model's decision must not significantly alter the very decision that we are trying to explain. Also, they must not utilize any reference other than the model itself to derive the explanations \cite{Lundberg2017}. Current keyframe extraction methods \cite{dong2022identifying} \cite{Ghodrati2021} \cite{Wu2019} are designed to enhance (hence change) the model decision. Some methods use an external model to find the keyframes \cite{dong2022identifying}. Such traits make existing methods of keyframe extraction unsuitable for explaining model decision. We introduce the first Model Conditioned Keyframe Extraction and Fair Attribution (MoCKEFA), an algorithm that coalesces a video stream into cohesive groups sharing a common core characteristic that is essential for the model decision, and extracts a representative keyframe from each group (Figure \ref{fig:intro}). Crucially, repeating this keyframe in place of the full group does not significantly alter the model's belief, revealing that intra-group changes are largely inconsequential to the decision. Instead, only transitions between keyframes carry meaningful temporal information. This refined understanding of temporal attribution promises to be critical in domains such as medical AI and autonomous driving \cite{meng2019interpretable}.

Keyframe attribution can be framed as a cooperative game-theoretic problem, where n players jointly contribute to a total payoff. The central question, then, is how to define a \textit{fair} distribution of this value. Lloyd Shapley \cite{Shapley1953} addressed this by proposing four axioms that any fair allocation must satisfy (see Section 2); their intuitive appeal has made them the de facto standard for fairness. Shapley further proved that the Shapley value is the unique solution satisfying these axioms, and it has since been widely adopted to compute individual payoffs in n-player games. In computer vision, this framework has been readily adapted—notably, Play Fair \cite{Price2020} employed Shapley values for video frame attribution. Following this lead, we treat each keyframe as a player, with the model's prediction serving as the total payoff.

\begin{figure*}
	\centering
	\includegraphics[width=1.0\linewidth]{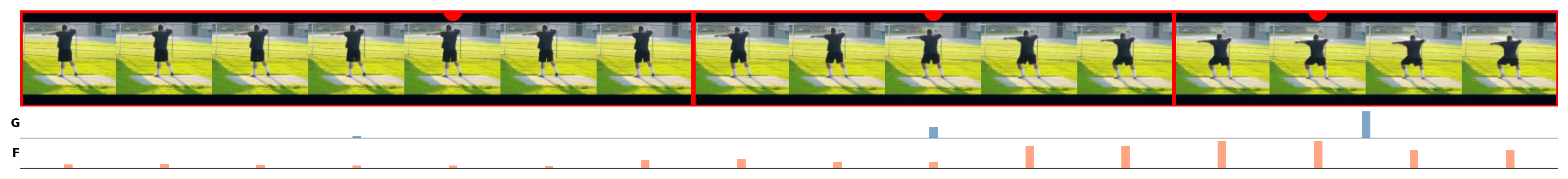}
	\caption{Comparison of frame-wise (F) and group-wise (G) attribution. Keyframes have a red dot on the top. The three moments represent three stages of the bending of the left leg of the person.}
	\vspace{-1.5em}
	\label{fig:frame_vs_grp}
\end{figure*}

In frame-wise Shapley attribution such as in Play Fair \cite{Price2020}, mutually redundant frames receive nearly equal payoffs—but these payoffs reflect only the unique information each frame contributes relative to others, not any intrinsic value embedded within the frame itself. Consequently, per-frame payoffs alone cannot reliably reveal the true importance of each core characteristic (main distinct concepts in the video). By assigning attribution at the keyframe level, we obtain a far more interpretable representation. Consider Figure \ref{fig:frame_vs_grp}, where the video is partitioned into three visually coherent shots. Plot G (keyframe-wise) and Plot F (frame-wise) tell contrasting stories: while most frames within a group receive similar frame-level scores, the human mind more naturally grasps the per-keyframe, per-core-characteristic assignment of attribution in plot G. Beyond interpretability, keyframes also offers a practical computational advantage—exact Shapley values require $2^{n\_players}$ model evaluations, so reducing the number of players dramatically lowers cost. That said, Figure \ref{fig:frame_vs_grp} reveals a subtle limitation: the last frame of the first group and the last two frames of the second group receive above-average attribution, as they lie near the boundary of the next group—which has higher average attribution. This suggests that MoCKEFA by discretizing a continuous physical process, introduces a mild approximation. Nevertheless, we find this approximation sufficiently accurate for our analysis, and in Section \ref{sec:eval} we show that finer groupings can arbitrarily approximate the original video.

In general, video understanding models cannot process variable video lengths. A problem arise when we have to remove certain frames in the video to estimate keyframe attributions. Play Fair \cite{Price2020} creates a new multi-scale model that can process variable lengths. But if we calculate attribution with such a secondary model we are actually estimating the attribution given by this new model, not the original model. On the other hand our attribution method does not require any modifications to the model and will work on any model architecture. We show that temporal freezing can be used to significantly reduce the burden of evaluating on a background set required by Shapley value calculation. Temporal freezing also gives more intuitive and quantitatively better results than other filling methods such as all zero values and sampling from a distribution of other videos. 

Our contributions to video temporal attribution:
\begin{itemize}
	\item We introduce the first model conditioned keyframe extraction algorithm suitable for explaining video models that can 1) Use the model output to group adjacent frames of a video into several shots. 2) Extract a keyframe from each shot that can represent the information contained in each shot that is valuable to the model decision.
	\item Show that temporal freezing is more intuitive and qualitatively better than other methods to fill in missing frames for temporal attribution in videos.
	\item Show that Sahpley-based keyframe attributions makes more intuitive sense and easy to interpret than other frame-wise attribution baselines. 
\end{itemize}

\section{Related Work}
\label{sec:related}

\textbf{Selecting keyframes for inference} Oleksii et al \cite{Sidorov2020} observed strong aggreement among the participants on which spacial and temporal sections of a video are important for them.  This experiment indicates that temporal attribution is a critical dimension of explainable video understanding. FrameExit \cite{Ghodrati2021}, AdaFrame \cite{Wu2019} and Dong et al. \cite{dong2022identifying} can select a "good" set of frames for a reliable inference. Xia et al. \cite{Xia2022} showed that only a few frames in a video may be truly salient. While existing approaches like shot boundary detection \cite{zhu2023autoshot} and temporal segmentation \cite{ding2023temporal}\cite{wu2023newsnet} group similar frames using external cues, we require the model itself to determine which frames are mutually redundant.

\textbf{Filling missing frames} A common way to evaluate the importance of a feature is to evaluate the model on situations where the feature is absent. The feature must often be replaced with a neutral. The neutral value must not leak any information about the removed features to the model while not significantly changing the input distribution. So the choice of the neutral value is non-trivial. Researchers commonly use a constant value, noise and Gaussian blur as ways to fill in the missing parts of the image \cite{Fong2017} \cite{Sturmfels2020} \cite{Sturmfels2020}. Pixels of constant colors are commonly used \cite{Ribeiro2016} \cite{Zeiler2014} but might create out-of-distribution samples \cite{Sturmfels2020}. Replacing the missing values of data with sampled values from a background dataset has been a standard method specially in the Shapley value literature \cite{Strumbelj2010} \cite{Lundberg2017} \cite{Ribeiro2016}. Although this make sense for some less complex data modalities it will create out-of-distribution samples in highly comlex data modalities such as video \cite{Sundararajan2020}. Temporal freezing has been used to fill in removed video frames \cite{Manttari2020} \cite{Huang2018}.

\textbf{Feature attributions} Several feature attribution methods have been used for machine learning. Guided backpropagation \cite{Sundararajan2017} and GradCAM relies on the model gradients \cite{Selvaraju2017}. Integrated Gradients \cite{Sundararajan2020} has some favorable theoratical guerentees. LIME \cite{Ribeiro2016}, DeepLIFT tries to avoid some problems inherent to earlier methods \cite{Shrikumar2017} \cite{Shrikumar2016}. Occlusion Sensitivity Analysis (OSA) masks certain regions in the input to see if those areas are important for the model \cite{Zeiler2014}. Video attribution is more complicated than images because of the  spacio temporal relationships between various sections of the input video \cite{Li2021}. Some examples are as follows. GradCAM++ \cite{Chattopadhay2018}, Saliency tube \cite{Stergiou2019} \cite{Zhou2016}, SWAG-V \cite{Hartley2022} \cite{Hartley2021}, STEP \cite{Li2021} \cite{Fong2019}, Adaptive occlusion analysis for video \cite{Uchiyama2023}. Play Fair worked on the rarely studies topic of video frame attribution \cite{Price2020}. Shapley values are considered the gold standard in feature attributions because they are the only formulation that satisfy the following axioms. \textbf{Efficiency}: The individual payoffs of each player must add up to the total value. \textbf{Symmetry}: If two players contribute equally under all possible coalitions, their payoffs must be equal. \textbf{Dummy}: If a player does not change the total value under any coalition, it should have zero payoff. \textbf{Additivity}: If we create a new game with combined payouts of two other games, the payoff of each player must add up \cite{Shapley1953} \cite{Lundberg2017} \cite{young1985monotonic}. The exact calculation of Shapley values is very computationally heavy and requires $2^K$ number of model evaluations for $K$ number of features. There are methods that can be used to approximate Shapley values with less computational requirements. For example, Kernel SHAP \cite{Lundberg2017}. Shapley sampling values \cite{mitchell2022sampling}. Partition SHAP creates a hierarchy of features and calculates the Shapley values along this hierarchical tree. This both allows for inter feature correlations to be accounted and for faster computation \cite{xu2025succinct} \cite{matthews2021mshap} \cite{nayebi2023windowshap}. The hierarchical structure can be manually defined and hence allows us to define in a way that considers the temporal ans sequential nature of video data. Play Fair \cite{Price2020} applies Shapley values to solve the video frame attribution.

\section{Model Conditioned Keyframe Extraction}
\label{sec:formatting}

\begin{figure*}[t]
  \centering
   \includegraphics[width=\textwidth]{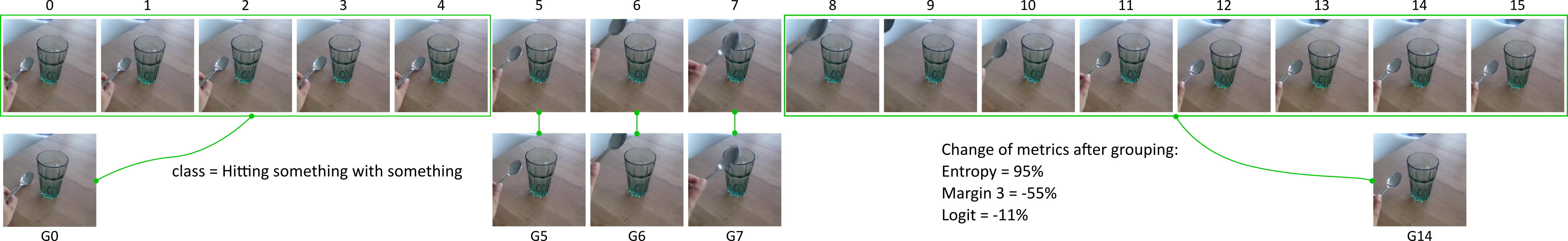}
   \caption{Example of a distilled video. Frames 0 to 4 can be represented just by the frame 0. Frames 5, 6 and 7 are groups with just one frame. Frame from 8 to 15 can be represented with just the frame 14. Temporal Freezing can be used to create the new video with the original length. Check Sec \ref{sec:kf_qual} for metric definitions}
   \vspace{-1em}
   \label{fig:grp_vid}
\end{figure*}

As adjacent frames of a video are often highly redundant, we can represent many sets of adjacent frames with just a single keyframe extracted from the same set. We introduce model conditioned keyframe extraction, which partitions a video into groups with the same core characteristic guided by the model and extracts a single representative keyframe from each group. Let's represent a video as an ordered sequence of frames as $V = [I_1, I_2, \ldots, I_T]$. Let the video be partitioned into $k$ groups $\mathcal{G} = \{G_1,G_2,\ldots,G_k\}$. Each group  $G_i$ contains one or more adjacent frames $G_i = \{I_t \mid s_i \leq t \leq e_i\}$ where $s_i$ and $e_i$ are the start and end frame indices. Each group $G_i$ has a representative keyframe $K_i$ that can be used to represent the whole group. The resulting distilled video is the ordered sequence $V_\diamondsuit = [K_1, K_2, \ldots, K_k]$ which contains the set of keyframes $\mathcal{K}$. An example of video distillation is shown in Figure \ref{fig:grp_vid}.

In a related work, Huang et al \cite{Huang2018} ordered the frames according to their predicted class logit response. The top n best frames were repeated (temporal freezing) to create a video of full length. They observed a significant drop in performance. The problem with this method is that the selected n frames might contain very similar information and the collection of frames is not necessarily better than using just one frame. In contrast our method can select keyframes that have independent value to the model. 

\textbf{The algorithm}: Consider frames 1 and 2. Find best frame from $\{1,2\}$ to replace both the frames (which choice degrades the model belief the least). This frame is the current keyframe $kf$. Consider frame 3. Find the best frame from $\{kf,3\}$ to replace frames 1,2 and 3. Update $kf$. Keep doing this until the model belief degrades more than the threshold. When that happens finish distilling group 1 and repeat the above process starting with the next frame that does not already belong to a group to find the next group.

\begin{algorithm}
    \caption{Model-guided keyframe extraction algorithm. [i,j]: all elements from i to j including both}
    \begin{algorithmic}[1]
    	\Require Video $vid$, Threshold $THR$, model $model$
    	
    	\Function{best\_frame}{vid, [grp], i, j}
    		\State $ v  = $ \Call{Copy}{vid}
    		\State $ v[grp + i] = v[j]$
    		\State $ \Delta_{mj} = \Call{margin\_change}{model, v}$
    		
    		\State $ v =$ \Call{Copy}{vid}
    		\State $ v[grp + j] = v[i] $
    		\State $ \Delta_{mi} = \Call{margin\_change}{model, v}$
    		
    		\State \Return $(i,\Delta_{mi}) $ or $(j,\Delta_{mj})$ with min $\Delta_{m}$

    	\EndFunction
    	
        \Function{extract\_keypoints}{vid, THR} \Comment{Main}
            \State $T=$ number of frames in $vid $ 
            
            \State $i = 0$, $d = \{\}$
            \While{$i < T$}
            	\State $i^*,\Delta_{mi^*} = \Call{best\_frame}{vid,[],i,i+1}$
                
                \State j = i+1
                
                \While{ $\Delta_{mi^*} <$ THR}
                	\State j = j+1
                	\State $i^*,\Delta_{mi^*}$=\Call{best frame}{$vid,[i,j-1], i^*,j$}
                \EndWhile
                \State Append group: $(i^*, [1,j]) \rightarrow$ d
                \State $i \gets$ j+1
            \EndWhile
            \State \Return d
        \EndFunction
    \end{algorithmic}
\end{algorithm}
\noindent

\textbf{Important changes in the video}: A keyframe alone can represent a group of frames means all the changes that the model value are located between adjacent groups. Changes withing the same group are insignificant to the model. We can estimate the relative importance of each of these changes in future work.

\textbf{Why independent grouping?} According to the Algorithm 1, when creating a new group we start with a fresh video with no distillation. When creating group 2, we do not use the video with group 1 already distilled (all the frames in the group 1 replaced with the keyframe). This is because we need to check how the model belief change when we form the group 2 independent of group 1. If we start with group 1 already distilled, the group 2 we find will be conditioned on group 1. We want all the groups to capture the similar regions of the video in an equal quality and strength. But this independent grouping creates a secondary problem. We cannot directly control the quality of the distilled video. We can only measure the final quality of the distilled video once all the groups are being found and distilled. We show that we can control the final quality of the grouped video by changing the grouping threshold. Higher threshold results in coarse groups where the video will be represented by only a few groups. While the brevity of the representation will increase, the potential for the video to deviate from the original one is higher. Lower threshold will result in finer groups where the video will be represented with a greater number of groups. In this case the grouped video will be more similar to the original one, but the brevity of the video will decrease.

\section{Temporal Keyframe Attribution}

We estimate keyframe attribution for the distilled video using Shapley values \cite{Shapley1953} \cite{Lundberg2017}. When calculating the Shapely values we get the difference between the model output with and without a given keyframe under all the possible combinations of other keyframes. Since a general video understanding model cannot process videos of arbitrary length, we need to fill in missing frames with a suitable fill-in frames after removing a keyframe. See section \ref{sec:filling} for a discussion on filling methods. We start with the distilled video $V_\diamondsuit$ which contains the keyframes in the set $\mathcal{K}$. To remove a keyframe $K_i$ from $V_\diamondsuit$, repeat the existing keyframes according to the lenght of the groups they belong and to fill in the missing values, we use a fill operator $\mathcal{F}(V_\diamondsuit, S) = V_\diamondsuit^{S}$ where $ S \subseteq \mathcal{K} $. We assume the full distilled and filled video is a proxy for the original video $V_\diamondsuit^{\mathcal{K}} \approx V$. 

\vspace{-1.4em}
\begin{equation}
	\phi_i(V_\diamondsuit){=}\hspace{-1em} \sum_{S \subseteq \mathcal{K} \setminus \{i\}} \hspace{-1em} \frac{|S|! (|\mathcal{K}| - |S| - 1)!}{|\mathcal{K}|!} [f(V_\diamondsuit^{S}) - f(V_\diamondsuit^{S \setminus i})]
	\label{eq:shap}
\end{equation}
\vspace{-0.7em}

\noindent where: $\mathcal{K}$: set of all keypoints, $S$: a subset of keypoints chosen from $\mathcal{K}$ which does not include the group $i$. $V_\diamondsuit^{S}$: distilled and filled video which only contains keyframes in $S$. Frames belonging to other groups have been removed and filled-in with a desired filling method. 

Following the common intuition, earlier work \cite{Price2020} estimated frame Shapley values. We on the other hand do keyframe attribution since they provide more intuitive explanations and reduced compute requirements (Section \ref{sec:intro}).

Video understanding models in general cannot process inputs with variable lengths. Play fair  \cite{Price2020} built a multi-scale model which can handle variable-length inputs. But training several models to build a multi-scale model is very compute and time consuming. Most importantly, it will estimate temporal attributions relative to the multi-scale model; not the original model. Our method has no limitations on model architectures, such as transformers \cite{arnab2021vivit} and LSTMs \cite{bin2018describing}, which can handle variable input sizes and makes no modifications to the model. To achieve this, we fill in missing frames with "neutral" frames to simulate the lack of frames. 

\subsection{Filling methods}
\label{sec:filling}
 The method used to fill in the missing frames must minimally change the temporal consistency of the video. Consider the example in Figure 1 for instance. After distillation, we are left with keyframes 2,3,5 and 7. Let's think about how we are going to remove keyframe 5. How will the video look if the events shown in keyframe 5 never happened, but events surrounding it which are 3 and 7, still happened? We explore several options next. 
     
Let's say we have a distilled video with keyframes 1,2,3,4,5, and 6. Each group consist of the keyframe repeating one or several times. In this example, group 1 consists of a keyframe repeating twice, group 2 ones, group 3 four times, group 5 three times and group 6 one time. Assume we would like to remove group 3 while keeping the number of frames in the video unchanged. There are an infinite number of ways to fill in the missing frames. We experimented with several cases while avoiding several methods that would clearly break the consistency of video.

\noindent
\[
\begin{array}{l l}
	v      & \texttt{1 1 2 3             3            3              3             4 5 5 5 6} \\
	v_r    & \texttt{1 1 2 \underline{\phantom{2}} \underline{\phantom{2}} \underline{\phantom{2}} \underline{\phantom{2}} 4 5 5 5 6}  \\
	past   & \texttt{1 1 2 \underline{2} \underline{2} \underline{2} \underline{2} 4 5 5 5 6}  \\
	future & \texttt{1 1 2 \underline{4} \underline{4} \underline{4} \underline{4} 4 5 5 5 6}  \\
	mid    & \texttt{1 1 2 \underline{2} \underline{2} \underline{4} \underline{4} 4 5 5 5 6}  \\
	random & \texttt{1 1 2 \underline{2} \underline{2} \underline{2} \underline{2} 4 5 5 5 6}  \\
		   & \texttt{1 1 2 \underline{2} \underline{2} \underline{2} \underline{4} 4 5 5 5 6}  \\
		   & \texttt{1 1 2 \underline{2} \underline{2} \underline{4} \underline{4} 4 5 5 5 6}  \\
		   & \texttt{1 1 2 \underline{2} \underline{4} \underline{4} \underline{4} 4 5 5 5 6}  \\
		   & \texttt{1 1 2 \underline{4} \underline{4} \underline{4} \underline{4} 4 5 5 5 6}  \\
\end{array}
\]
\noindent

\textbf{Past} fills the missing frames with the immediate frame from the past (left). If there are no frames present in the past, this will take the immediate future frames. \textbf{future} uses the immediate future frame. If there is no future frames, it uses the immediate past frame. \textbf{Mid} fills the left half of the missing frames from the past and the right half from the future. If there are an odd number of frames to be filled, a random choice is made between he two middle points. \textbf{Random} chooses a random number of frames from the number of frames to be filled and this number of frames are filled from the left to right with the immediate frame from the past. The rest is filled with the most immediate future frame. In all these cases, if there are no frames in the past, they are filled with the most immediate future frame and vise-versa. \textbf{Zero} fills all the missing frames with all-zero frames. \textbf{Late} is a combination of past and future methods. During the Shapley value calculation, we need to obtain the model inferences $f(V^S)$ where $S$ is a subset of the groups. Let $V_{p}^S$ be the video created with past method of filling while $V_{f}^S$ be the video created with the future method. Under the late method we calculate $f(V_{l}^S) = 0.5*(f(V_{p}^S) + f(V_{f}^S))$ and use $f(V_{l}^S)$ in Shapely value calculation.  

\textbf{Why not interpolate to generate the missing frames ?} Filling-in the missing frames of a video with interpolation is a well-established technique \cite{Ciubotariu_2026_CVPR}. We noticed that interpolation can inadvertently re-introduce the information from removed frames back into the video. For example assume we removed the frame 4 from the original video in Figure \ref{fig:intro}. A good interpolation tool can look at frame 3 and 5 and interpolate the location of the ball and the person's hand for frame 4 successfully. This will defeat our intention to remove the frame 4 in the first place. But shouldn't the model be able to interpolate the missing information anyway, at least in some cases? Yes, this may be true. If the model can actually regain the lost information using its internal knowledge, the model will not "think" the missing frames are important, and the missing frames will get a low importance score since when we remove the frame in question, the model belief will not change significantly.

\section{Evaluations}
\label{sec:eval}

\subsection{Models and Datasets}
We use the UCF101 dataset \cite{soomro2012ucf101} which contains 101 classes of different activities. All the evaluations are performed on the 3783 test samples of the split 1 as defined in \cite{kataoka2020would} and the R3D model fine-tuned on UCF-101 by \cite{Uchiyama2023}. We use the a subset of the test set of the SSV2 dataset \cite{goyal2017something}\cite{mahdisoltani2018effectiveness} and randomly sample 6 samples from each of the 174 classes to create a daaset of 1044 samples due to restrictions in computing resources. We use V-JEPA model \cite{assran2025v} which is a ViT-L 256 model with video classification head pretrained on the SSV2 dataset.     

\subsection{Keyframe extraction quality}
\label{sec:kf_qual}
We want to distill (keyframe extraction + filling) the video in a way that minimally affect the model's belief. This manner we can make sure that the distillation does not affect the downstream attribution task. A plot of the model feature distributions taken from the distilled vs original videos did not show any significant deviation (Appendix 2.a). The Tables \ref{tab:dist_ucf} and \ref{tab:dist_ssv} shows the change of various metrics after distillation. We use the following metrics to measure distillation quality. The $k^{th}$ margin is defined as $	m_k = z_{(1)} - \frac{1}{k} \sum_{i=2}^{k+1} z_{(i)}$ Let $z_{(1)} \ge z_{(2)} \ge \cdots \ge z_{(C)}$. This compares the best prediction logit to a background of $k$ next best logits. Entropy of a prediction is defines as $E(p) = -\sum_{i=1}^{C} p_i \ln p_i$. We are interested in how these metrics changes after distillation. We define the logit change as $\Delta_{L} = \frac{L_o - L_d}{L_o}$. This signifies how has the max logit changed after distillation ($L_d$) with respect to the original video ($L_o$). Similarly we also define the entroy change and margin change as $\Delta_{E} = \frac{E_o - E_d}{E_o}$ and $\Delta^{m^{k}} = \frac{m^{k}_o - m^{k}_d}{m^{k}_o}$. We used $\Delta^{m^{3}}$ as the metric when doing the distillation. We do not just use the logit because we want to track a more nuanced view of model belief than provided by the simple logit. In certain cases the inference class may change after distillation. The agreement, $Ag\%$ measure the percentage of samples in the dataset that did not change their prediction class after distillation. $\mu_k$ is the average number of keyframes found. 

From Tables \ref{tab:dist_ucf} and \ref{tab:dist_ssv}, it is evident that as we lower the threshold, more keyframes are found (increasing $\mu_k$) and the $Ag\%$ increases. This is because lower thresholds allow more nuanced and milder distillation which resembles the original video more closely. Negative values for $\Delta_{L}$ and $\Delta_{m}$ indicates that the model's confidence of its original prediction increased after distillation compared to the original video. As we decrease the threshold the model confidence on average increased compared to the original video (also evidenced by increasing $\Delta_{entr}$) and finally it decreases and comes back to the original level. This is because at first the keyframe extraction algorithm finds better keyframes that boosts the model decision (by removing noisy/unnecessary frames). But as the distillation comes closer to the original video this advantage vanishes. Although tempting, increasing the model confidence does not always results in better attributions as shown in Sec \ref{sec:exp_dist_vs_quality}. We observe that $\mu_k$ for the same threshold is higher for SSV2 compared to UCF-101. This may be due to SSV2 having more motion-centric videos where minute detailes in motion is more important than UCF-101.

\begin{table}
	\centering
	\setlength{\tabcolsep}{2pt} 
	\begin{tabular}{@{\extracolsep{\fill}} r r r r r r r r }
		\toprule
		Thr & $\mu_k$ & $\Delta^{L}$ & $\Delta^{m1}$ & $\Delta^{m3}$ & $\Delta^{m5}$ & $\Delta^{entr}$ & $Ag\%$ \\
		\midrule
		-1e-1 & 14.98 & -0.011 & -0.587 & -0.078 & -0.051 & 0.043  & 99.55\\
		-1e-2 &  7.01 & -0.024 & -1.327 & -0.131 & -0.089 & 0.119  & 98.65\\
		-1e-3 &  4.72 & -0.016 & -1.408 & -0.118 & -0.076 & 0.017  & 98.28\\
		-1e-5 &  4.51 & -0.015 & -1.406 & -0.116 & -0.074 & -0.020  & 98.31\\
		1e-4 &  4.48 & -0.022 & -1.410 & -0.116 & -0.074 & -0.021  & 98.28\\
		5e-4 &  4.41 & -0.014 & -1.407 & -0.115 & -0.072 & -0.037  & 98.26\\
		1e-3 &  4.33 & -0.013 & -1.381 & -0.114 & -0.072 & -0.046  & 98.23\\
		5e-3 &  3.96 & -0.009 & -1.380 & -0.105 & -0.064 & -0.098  & 98.10\\
		1e-2 &  3.66 & -0.005 & -1.391 & -0.096 & -0.055 & -0.166  & 97.96\\
		\bottomrule
	\end{tabular}
	\caption{UCF101 Metrics after distillation}
	\label{tab:dist_ucf}
\end{table}

\begin{table}
	\centering
	\setlength{\tabcolsep}{2pt} 
	\begin{tabular}{@{\extracolsep{\fill}} r r r r r r r r }
		\toprule
		Thr & $\mu_k$ & $\Delta^{L}$ & $\Delta^{m1}$ & $\Delta^{m3}$ & $\Delta^{m5}$ & $\Delta^{entr}$ & $Ag\%$ \\
		\midrule
		-1 & 15.89 & -0.006   & -0.183 & -0.077 & -0.043 & 0.012 & 99.81 \\
		-1e-1 & 11.87 & -0.031 & -3.690 & -0.272 & -0.187 & 0.170 & 96.65 \\
		1e-5 & 6.45 & 0.039 & -6.608 & -0.116 & -0.054 & -0.566 & 90.52 \\
		1e-4 & 6.44 & 0.039 & -6.606 & -0.115 & -0.053 & -0.570 & 90.52 \\
		5e-4 & 6.45 & 0.038 & -6.644 & -0.117 & -0.054 & -0.566 & 90.61 \\
		1e-3 & 6.39 & 0.040 & -6.641 & -0.113 & -0.051 & -0.577 & 90.52 \\
		5e-3 & 6.19 & 0.043 & -6.562 & -0.102 & -0.043 & -0.643 & 89.94 \\
		1e-2 & 5.93 & 0.051 & -6.518 & -0.083 & -0.029 & -0.960 & 89.37 \\
		\bottomrule
	\end{tabular}
	\caption{SSV2 Metrics after distillation}
	\label{tab:dist_ssv}
\end{table}

\begin{figure*}[t]
    \begin{subfigure}[b]{1.0\textwidth}
		\centering
		\includegraphics[width=\textwidth]{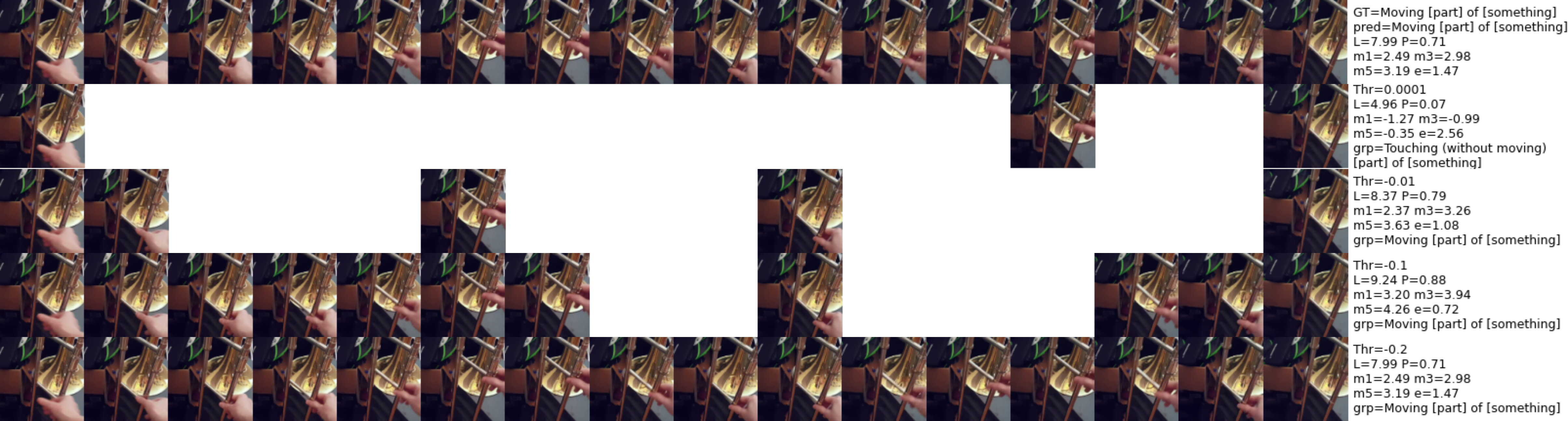}
	\end{subfigure}

	\caption{Iterative distillation till the desired similarity to the original video is achieved. Top figure is the original video. Then from top to bottom we decrease the threshold to distill with more finer groups.}
	\vspace{-1.5em}
	\label{fig:itr_crys}

\end{figure*}

\textbf{Iterative distillation} We can iteratively change the threshold till we get the desired similarity of the distilled and the original videos. See Figure \ref{fig:itr_crys} for an example. In this manner we can arbitrarily increase the similarity between the distilled and the original videos. But we use a fixed threshold for our attribution experiments due to simplicity.     

\subsection{Filling methods}
\begin{table}
	\centering
	\begin{tabular}{@{}c c c c c@{}}
		\toprule
		\multirow{2}{*}{Method} & \multicolumn{2}{c}{UCF-101} & \multicolumn{2}{c}{SSV2}\\
		\cmidrule(lr){2-3} \cmidrule(lr){4-5}
		& L2 & KL-Div & L2 & KL-Div\\
		\midrule
		future & 0.00875 & 0.04164 & 0.01168 & 0.01321\\
		past   & 0.00877 & 0.04182 & 0.01174 & 0.01396\\
		late   & 0.00867 & 0.04116 & 0.01175 & 0.01400\\
		mid    & 0.00872 & 0.04137 & 0.01174 & 0.01369\\
		random & 0.00870 & 0.04124 & 0.01169 & 0.01354\\
		zero   & 0.01212 & 0.08637 & 0.01425 & 0.01800\\
		\bottomrule
	\end{tabular}
	\caption{We removed a single randomly chosen group from each crystallized video and filled in the missing values using different methods. Then we obtained the model embeddings of these filled videos. This table shows the difference in the distributions of the original embeddings and the filled embeddings. L2: Average sample-wise Euclidian distance. KL-DIV: Average sample-wise KL divergence}
	\label{tab:filling}
\end{table}

To investigate which type of filling method is the most suitable, we randomly removed a group from distilled videos from the two datasets and filled it out with different filling methods. We calculated the L2 distance and the KL-Divergence between the original and the filled video embeddings taken from the model. From Table \ref{tab:filling} it can be seen that the $late$ is the best method for UCF101 and while $future$ is the best for SSV2. Further all the temporal freezing based filling methods produced distributions that are very close to the original while filling with all-zero frames produced a distribution that is more different. The small discrepancy between the two datasets might indicate that different types of videos may require different types of filling methods. We might be able to estimate better temporal attribution if we used the optimum filling type for each video rather than the dataset best for all the individual videos. This in an interesting direction for further investigation in the future. 

\subsection{Attribution Results}
\textbf{Metrics}:
To evaluate the validity of the temporal attributions, we conduct a systematic removal test on the distilled videos and its corresponding importance values. Starting from the full distilled video, we iteratively remove one keyframe at a time, fill-in the resulting gaps, and record the model's output performance. We use the prediction logit and the probability (logit after softmax) normalized for each video as metrics when calculating the AUC. The order of removal is determined by five distinct strategies: random ($AUC_{\text{ran}}$), ascending ($AUC_{\text{asc}}$), descending ($AUC_{\text{des}}$), left-to-right ($AUC_{\text{lr}}$), and right-to-left ($AUC_{\text{rl}}$). This design allows us to assess how the model's predictions degrade as frames are progressively withheld. We omit the addition test metrics in the main text, as they are conceptually redundant and yield consistent conclusions (Appendix 4.b).

\noindent \textbf{Group-wise attribution}:
We calculate all Shapley-based attributions using SHAP python package \cite{NIPS2017_7062}. We use three different stratergies to calculate Shapley values. \textbf{Exact}: Strictly follows equation \ref{eq:shap} and requires $2^{n\_grps}$ model evaluations. Calculated only for the UCF-101 dataset due to high compute requirements under the larger model we used for the SSV2 dataset. We use two methods that were design to reduce the computational requirements of exact Shapley calculation by means of approximation. \textbf{Partition}: Calculate Shapley values along a user-defined hierarchical structure of the groups \cite{Lundberg2017}. We defined an intuitive temporally-constrained hierarchy suitable for videos (Appendix 4.a). \textbf{Kernel}: Use linear regression \cite{Lundberg2017} and only used for UCF-101 since it provided very similar results to the Partition method.

\noindent \textbf{Frame-wise attribution}:
GradCAM \cite{Selvaraju2017} is used as it is a classic gradient based method. Integrated Gradients \cite{Sundararajan2017} is used as another axiomatic method. Occlusion \cite{Zeiler2014} is used as a perturbation method. We also calculate \textit{frame-wise} Shapley values using the partition method. All frame-wise methods gives per-frame attributions and we take group-wise mean to get per-keypoint attributions. 

Tables \ref{tab:attrib_p_ucf} to \ref{tab:attrib_l_ssv2} shows the removal test results on the two datasets. \textit{Each} group-wise attributions outperform \textit{each} frame-wise attributions. All three Shapley based methods for group-wise attribution for the UCF-101 yielded similar results showing that approximations found solutions close to the exact Shapley values (Tables \ref{tab:attrib_p_ucf} and \ref{tab:attrib_l_ucf}). For SSV2 datast, group-wise calculation of Shapley values with partition method shows the best performance again showing the superiority of group-wise attribution with Shapley values (Tables \ref{tab:attrib_p_ssv2} and \ref{tab:attrib_l_ssv2}). 

\begin{figure*}
	\centering
	\begin{subfigure}[b]{1.0\textwidth}
		\centering
		\includegraphics[width=\linewidth]{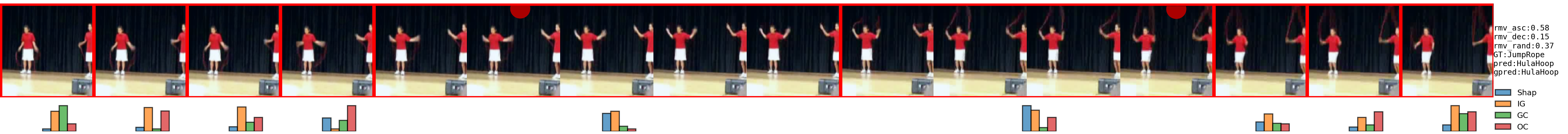}
	\end{subfigure}
	
	\vfill
	
	\begin{subfigure}[b]{1.0\textwidth}
		\centering
		\includegraphics[width=\linewidth]{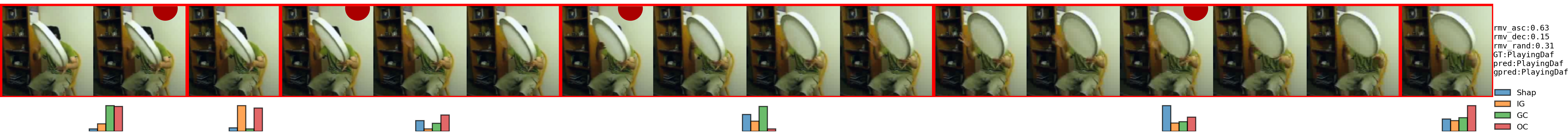}
	\end{subfigure}
	\vspace{-1.5em}
	\caption{Attribution examples - UCF101. Keyframes are shown with red a circle}

\label{fig:imp_ucf}	
\end{figure*}

\begin{figure*}
	\centering
	\begin{subfigure}[b]{1.0\textwidth}
		\centering
		\includegraphics[width=\linewidth]{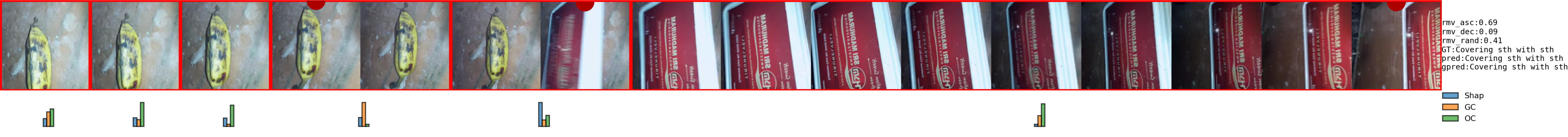}
	\end{subfigure}
	
	\vfill
	
	\begin{subfigure}[b]{1.0\textwidth}
		\centering
		\includegraphics[width=\linewidth]{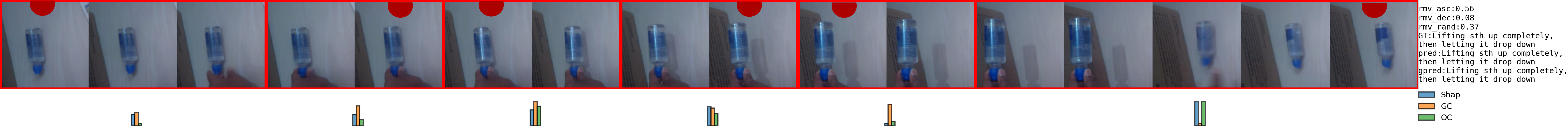}
	\end{subfigure}
	
	\vfill
	
	\begin{subfigure}[b]{1.0\textwidth}
		\centering
		\includegraphics[width=\linewidth]{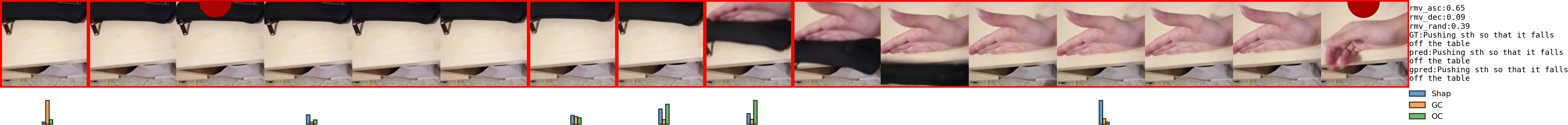}
	\end{subfigure}
	\vspace{-1.5em}
	\caption{Attribution examples - SSV2. Keyframes are shown with red a circle}
	
\label{fig:imp_ssv2}
\end{figure*}

Figures \ref{fig:imp_ucf} and  \ref{fig:imp_ssv2} show how the keyframe extraction finds groups of frames that can be grouped together and represented with a single keyframe. Most of the groups contain visually similar frames or contain redundant changes. For example in the middle part of Figure \ref{fig:imp_ssv2}, the first three frames where no motion happens is grouped together. The other sparsely-sampled keyframes may be sufficient for the model to understand the movements of the objects. Sometimes very similar adjacent frames are being put into different groups. We believe we will be able to solve these cases with iterative distillation. 

\setlength{\tabcolsep}{1pt}

\begin{table}
	\centering
	\caption{Removal test for the probability AUC of group attribution - UCF101. F- frame-wise attribution. G-Group-wise attribution}
	\begin{tabular}{@{}lccccccc@{}}
		\toprule
		&Method & $AUC_{asc} \uparrow $ & $AUC_{des} \downarrow $ & $AUC_{rand}$ & $AUC_{lr}$ & $AUC_{rl}$\\
		\midrule
		\multirow{4}{*}{F} & Partition & 0.708  & 0.663 & 0.688 & 0.674 & 0.697 \\
		&IG        &  0.690 & 0.688 & 0.688 & 0.674 & 0.697   \\
		&Gradam     & 0.686  & 0.692 & 0.688 & 0.674 & 0.697   \\
		&Occlusion  &  0.693 & 0.681 & 0.689 & 0.674 & 0.697  \\
		\midrule
		\multirow{3}{*}{G} & 	Kernel   &  0.722 & 0.650 & 0.688 & 0.674 & 0.697  \\
		&Partition  & 0.725  & 0.646 & 0.688& 0.674 & 0.697   \\
		&Exact    &  0.731 & 0.638 & 0.690 & 0.674 & 0.697  \\
		\bottomrule
	\end{tabular} 
	\label{tab:attrib_p_ucf}  
\end{table}

\begin{table}
	\centering
	\caption{Removal test for the Logit AUC of group attribution - UCF101. F- frame-wise attribution. G-Group-wise attribution}
	\begin{tabular}{@{}lcccccc@{}}
		\toprule
		&Method & $AUC_{asc} \uparrow $ & $AUC_{des} \downarrow $ & $AUC_{rand}$ & $AUC_{lr}$ & $AUC_{rl}$ \\
		\midrule
		\multirow{4}{*}{F} & Partition & 0.743  & 0.706 & 0.725 & 0.717 & 0.732  \\
		&IG        &  0.726 &  0.726 & 0.726 & 0.717 & 0.732   \\
		&Gradam     & 0.723  & 0.729 & 0.727 & 0.717 & 0.732   \\
		&Occlusion  & 0.730  & 0.720 & 0.726 & 0.717 & 0.732   \\
		\midrule
		\multirow{3}{*}{G} & Kernel   & 0.752  &0.697  &0.726 & 0.717& 0.732   \\
		&Partition  & 0.755  & 0.693 & 0.726 & 0.717 & 0.732   \\
		&Exact    & 0.758  & 0.688 & 0.727 & 0.717 & 0.732   \\
		\bottomrule
	\end{tabular}   
	\label{tab:attrib_l_ucf}  
\end{table}

\begin{table}
	\centering
	\caption{Removal test for the probability AUC of group attribution - SSV2. F- frame-wise attribution. G-Group-wise attribution}
	\begin{tabular}{@{}lccccccccccc@{}}
		\toprule
		&Method & $AUC_{asc} \uparrow $ & $AUC_{des} \downarrow $ & $AUC_{rand}$ & $AUC_{lr}$ & $AUC_{rl}$  \\
		\midrule
		\multirow{3}{*}{F} & Partition &  0.516 &0.380  & 0.462 & 0.441 & 0.402  \\
		&Gradam    & 0.471 & 0.477 & 0.463 & 0.441 & 0.402  \\
		&Occlusion & 0.471 & 0.439 &0.463  &0.441 & 0.402 \\
		\midrule
		\multirow{1}{*}{G} & Partition & 0.550 &0.340  & 0.460 & 0.441& 0.402   \\
		\bottomrule
	\end{tabular}   
	\label{tab:attrib_p_ssv2}  
\end{table}

\begin{table}
	\centering
	\caption{Removal test for the logit AUC of group attribution - SSV2. F- frame-wise attribution. G-Group-wise attribution}
	\begin{tabular}{@{}lccccccccccc@{}}
		\toprule
		&Method & $AUC_{asc} \uparrow $ & $AUC_{des} \downarrow $ & $AUC_{rand}$ & $AUC_{lr}$ & $AUC_{rl}$ \\
		\midrule
		\multirow{3}{*}{F} & Partition & 0.621 & 0.505 & 0.577 & 0.549 & 0.529   \\
		&Gradam    & 0.584 & 0.585 & 0.578 & 0.549 & 0.529  \\
		&Occlusion & 0.586 & 0.556 & 0.576 & 0.549& 0.529 \\
		\midrule
		\multirow{1}{*}{G} & Partition & 0.646  &0.470  & 0.576 &0.549 & 0.529 \\
		\bottomrule
	\end{tabular}   
	\label{tab:attrib_l_ssv2}  
\end{table}

\subsection{Distillation affects attribution quality}
\label{sec:exp_dist_vs_quality}
\begin{figure}
	\centering
	\includegraphics[width=0.8\linewidth]{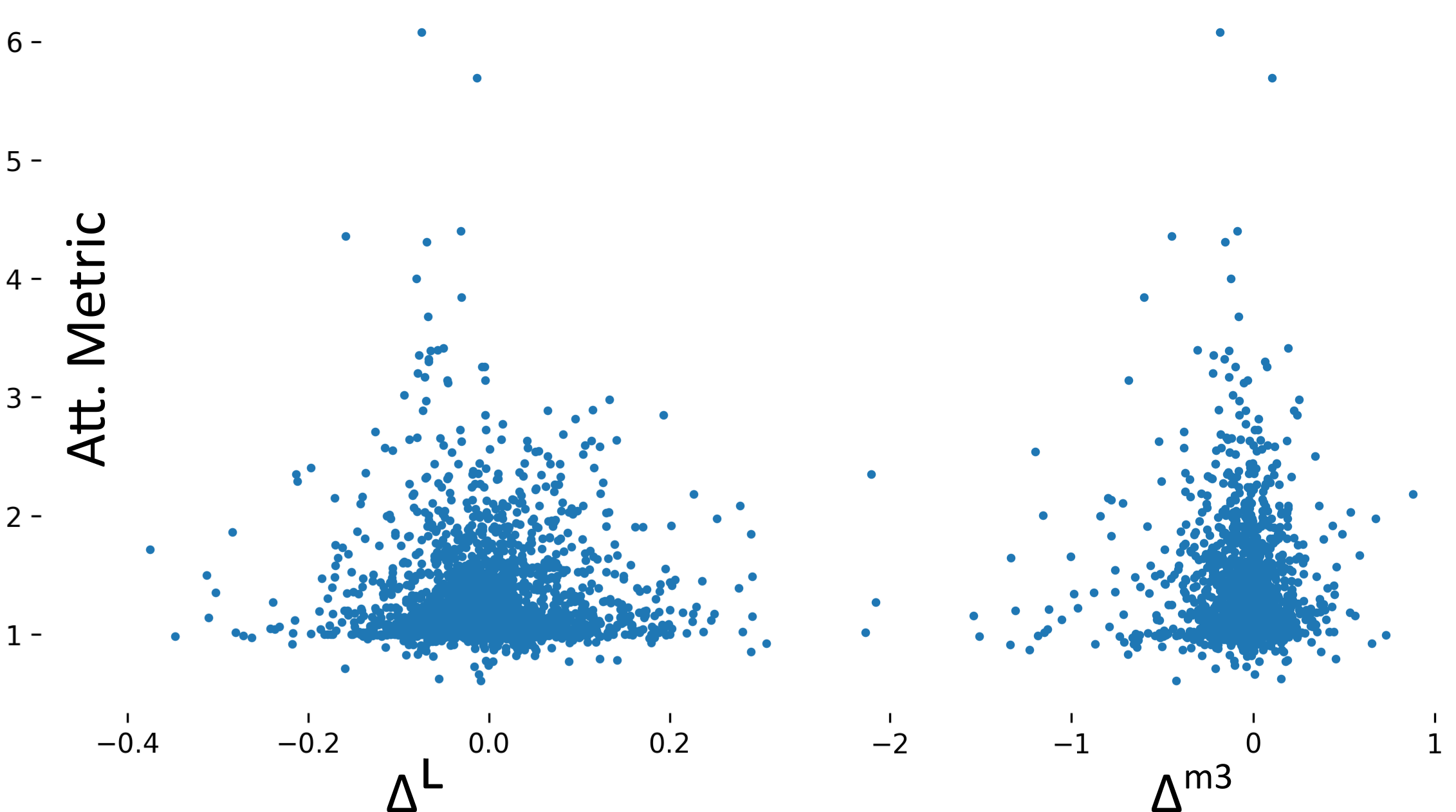}
	\caption{Effect of distillation quality on attribution}
	\label{fig:attmet_vs_grp}
\end{figure}

We create a single metric to measure the attribution quality by combining two removal AUC metrics $Att.Metric = \frac{AUC_{asc}^{R}}{AUC_{des}^{R}}$. Higher $Att.Metric$ indicates higher quality of attribution. Figure \ref{fig:attmet_vs_grp} shows the relationship between the attribution quality and the distillation quality. The keyframe attribution algorithm can do a better job when the distilled video is closer to the original video ($\Delta$ values closer to zero). Negative $\Delta$ values indicate that the model confidence about the predicted class increased after distillation. On average, the attribution quality degraded when this happens. It may be due to the data distribution being different from the original video. When the changed video further modified in the attribution algorithm, the distribution change may get amplified further. 

\subsection{Exploring important changes between frames}
\begin{figure}
	\centering
	\includegraphics[width=0.65\linewidth]{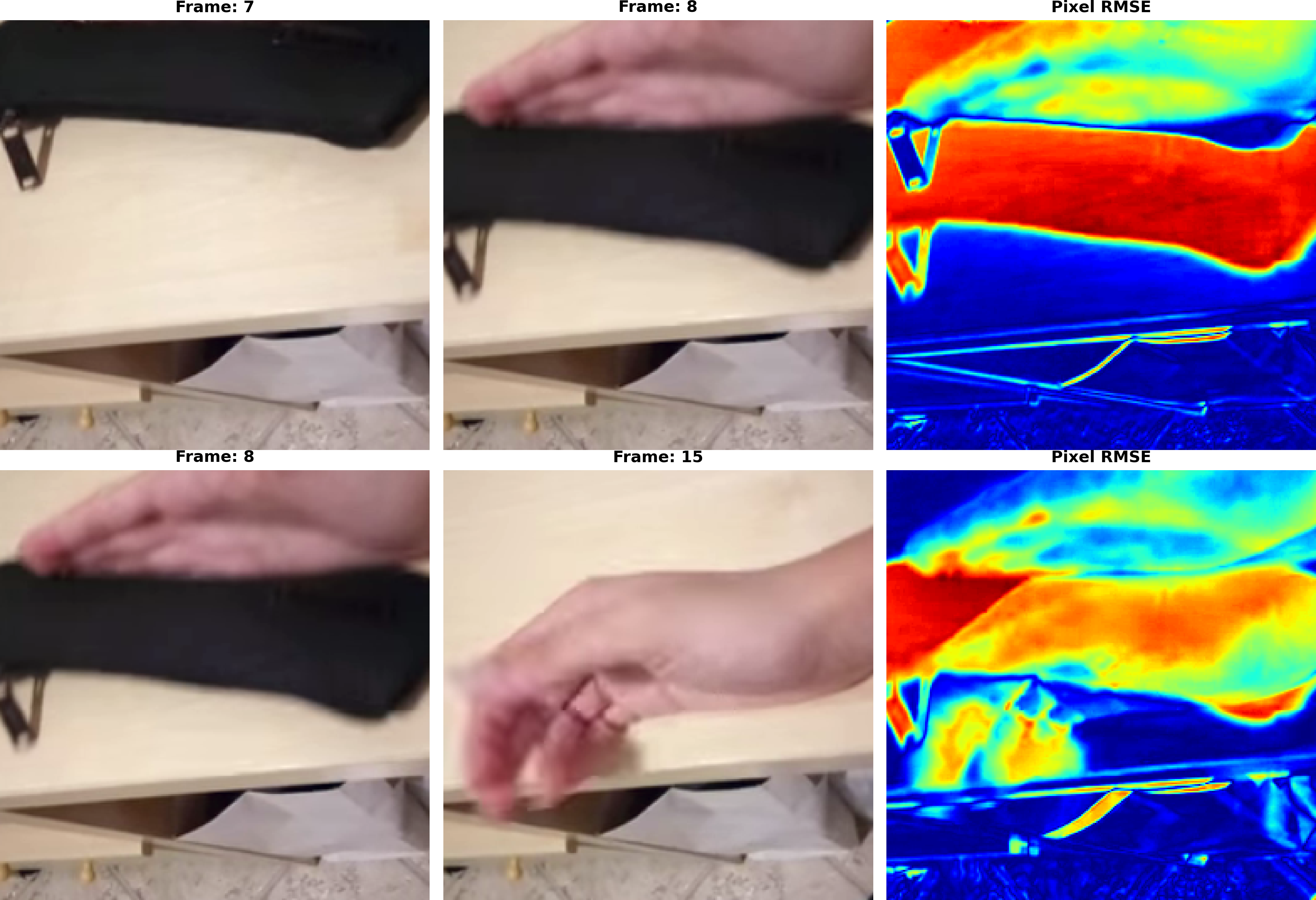}
	\caption{Important changes}
	\label{fig:motion}
\end{figure}

We can visualize important changes of the video by taking the pixel difference between adjacent keyframes. Refer Sec \ref{sec:intro} for a further discussion. Figure \ref{fig:motion} shows four keyframes from the same example as Figure \ref{fig:imp_ssv2} bottom part. RMSE between the images highlight the hand and the dark object that move between the keyframes.

\subsection{Conclusions}
MoCKEFA introduces the first video model explanation method that extracts keyframes from a video guided by the model and performs keyframe attribution. We show that the keyframe attributions are more intuitive and of higher quality than the baselines through qualitative and quantitative studies. Our use of fixed thresholds when extracting keyframes doesn't work for some examples and requires adaptive threshold selection. Also our distillation method is very compute intensive. These must be solved in future work.

\clearpage 
{
    \small
    \bibliographystyle{ieeenat_fullname}
    \bibliography{references}
}

\end{document}